# Controlling Traffic with Humanoid Social Robot

Faisal Ghaffar
Systems Design Engineering
University of Waterloo
Waterloo, Canada
faisal.ghaffar@uwaterloo.ca

*Abstract*—The advancement of technology such as artificial intelligence, machine learning and internet of things it became easy to develop more humanoid robots and automate different processes. An interactive robot must have high social behavior so that it can be easily accepted by the people using it. In this study we designed a traffic police robot (TRAPROB) to automate the traffic control at intersection. The human police officer experiences high stress because of long duty hours as well as pose the risk of accidents. The digital electronic signals are automatic but we want to create a system which is more human like and looks like an officer controlling the traffic at intersection. We used Thiago++ robot in this study and modified its look to like a police officer, and then programmed it to imitate and make gestures just like traffic police officer makes gestures for controlling traffic. We evaluated the looks, gestures, functionality, and social behavior of the robot. We asked a limited sample of two participants to identify the TRAPBOT, rate its look, the social behaviors and gestures in comparison to a real life police officer. we found that people can identify the robot as traffic police robot. Our analysis also shows that TRAPBOT has appearance like a traffic robot and can make similar signal gestures as a traffic police officer.

## I. Introduction

Human-Robot Interaction is a multidisciplinary field that deals with the development of robots for human use [1]. Over the last decade a wide variety of interacting robots have been developed. These robots have been developed in various fields such as healthcare, education, transportation and home automation etc. In this study our focus is to design and model a tra ffic underline police robot (TRAPROB) which assist the traffic police department in managing the traffic flow at intersection. To overcome the human resource limitation in traffic police department, accident and health risks of police personal, and control the digital traffic violations, it is imperative to design an social robots which can perform the traffic police duties.

Driving is one of the modern skills in this century. To meet the transportation needs, inter-city and intra-city road networks have been created. With the passage of time the number of vehicles and population has increased enormously and to meet these the smooth traffic requirement the roads length and widths has increased and the network became denser. To manage traffic the department of traffic police was organized in all major cities of the world. The first traffic police department was created in 1667 in the city of Paris [2].

The department evolved with time and can be seen in most modern form with access to latest technological equipment's. When the roads network and vehicles increased, the police personal were posted at different parts of the road to keep the traffic in flow and avoid any possible accidents. The major part of the road where traffic necessarily needs to be managed are intersections. A study carried out by Ghamdi et.al [3] in capital city of Saudi Arabia reported that almost 50% of accidents occurs at intersections.

Deploying traffic police personals has never been a feasible option. Although many strategies have been developed for deployment of police personal in the city [4] [5] [6] the human resource available with police department is always less sufficient to meet the needs of metropolitan city. These police personal works for longer hours and experience high stress level. Moreover because of longer duty hours in a congested traffic environment they are expose to high level of pollution [7] [8]. An alternative to traffic police personnel is to the use of electronic traffic control signal system. Most of the developing countries experience difficulties in deploying such system due to its high cost and people education and awareness. Conventional system works in a round robin way and it lead to vehicular delays and results in traffic congestion when traffic from one side is higher than the other [9]. Also these traffic signals are not human engaging and some studies shows people often lose interest in such systems [10] and as a result people often violate traffic signals. These traffic violation are much higher in an intersection where traffic signals are installed than at intersection where police personal are deployed.

To overcome these shortcomings we are proposing a humanoid robotic traffic management system (TROMS) to replace the traditional traffic police system and electronic signal installed at intersections. The main motivation behind these projects is to decrease the burden on traffic police department in metropolitan cities and replace the current system with social capabilities robot to engage drivers just like a police officer. The TROMS is based on a humanoid social robot which exhibits all the characteristics of a police personal and electronic signal system combined. The social robot with its robotic arms will direct the traffic in the same way as a traffic police personal and will also display the traffic signals accordingly. The hand, head and body movement of the robot provides a sense of police officer. A police personal uses arm with gestures such as extending arm in upward direction, lower direction, bending it half way, extending to the right direction and left direction. Combining different gestures with both hands create a traffic signal which directs vehicle from a certain direction to either stop or go. Our traffic robot works in the same way where we created different traffic signals from arms gestures. The robot head also imitates the police officer head and it can move in different directions along with the arm signals. The camera eyes and sensors installed in the robot reads the traffic flow from different directions and instead of acting in round robin fashion it will prioritise the heavy traffic road just like a human police officer. The humanoid social robot has two modes, it can either be controlled remotely from a control room or it can work autonomously. In this study we only focused on wizard of oz method for controlling the TRAPBOT. The robot social behavior was evaluated by two participants. In our experiments we found out that the RAPEBOT is identified as traffic controlling robot only when its installed in an traffic environment and it has police officer like appearance. The participant rated the the appearance of the robot is similar to police appearance with a high degree. The technical capability, gesture making ability, the interactive and social behavior were also evaluated and we found that the robot is highly suitable for the environment and can make clear, readable signals just like a police officer. We further find out through the participants that the speed of arms movement needs to be adjusted for the real environment. s

The rest of the paper is structured as follows The Sec. (II) discusses the related literature review and outlines some of the gaps. Sec. (III) explains about the robot structure Sec. (IV) explains various gestures made by the police personnel's. Sec. (V) explains our simulation methodology, robots behavior and one to one correspondence with police gestures. Sec. (VI) presents the experimental setup and the experiments which we carried out. Sec. (VII) discusses the overall simulation in light of of the experiments. Finally in Sec. (VIII) we conclude our study.

## II. LITERATURE REVIEW

Many studies have been carried out to design, model, implement, analyze interacting robots and also study their social behavior. In the field of healthcare, social assistance robots have great potential to ease the work of health care workers and assist the patients. It improve rehabilitation and health, and make the treatment process more enjoyable [11]. Wada et.al [12] [13] in their studies designed a robot Paro for therapeutic purposes. The robot has a reactive layer which responds to touch, sounds and light. The robot is also recognizes certain set of words and also respond to new frequently use words. Social robots have also been used for the therapy of autism patient. Dautenhahn et. al [14] carried out to study to explore the possibilities of using interactive virtual environments in the context of autism therapy. In another study by Pollack et. al [15] developed mobile robotic assistant Pearl for the old age people to remind people about daily activities such as eating, drinking, taking medicine, and guide them in their living place.

In the field of education interactive robots have been developed to help the children in learning. One such study is carried out by Kanda et al. [16]. They performed a study for two weeks with students and interactive humanoid robots. The humanoid robot acted like an english tutor and interacted with student by calling their names and also had a 50 words bank. later on in another version they added more step by step functionalities which would open up with more interaction. A similar longitudinal studies has been carried out by Tanaka et al. [17] and Kozima et al. [18]. Tanak et.al used QRIO robot with interactive behaviors such as choreographed dance sequences and imitating some of the toddler's movements. They reported that the interaction of toddler increased with time. Kozima et. al used Keepon robot which had non-verbal behaviours such as eye contact, joint attention and emotions. They reported in their study with passage of time children understand that instead of moving thing its more like a social agent.

Many domestic interactive social robot have also been developed to help in household activities. researchers have carried out studies to study the social behavior of those robots. Sung et al. [19] evaluated how people interact with robot vacuum cleaner 'Roomba'. They studied that acceptance and use of Roomba robots in their environment. Their experiment span was about six months during which an experimenter visited the 30 households five times. Based on their study they proposed four temporal steps that are pre-adoption, adoption, adaptation, and use and retention that explained the how the interaction patterns varies during different period of time. In a similar study of different age group, Klamer et. al [20] carried a study to find out how old age people use social robots at home. They programmed a Nabaztag robot to find out the factor that matters most in building relationship with a robot. They in their study interviewed the participants about the general use to find out the usefulness and the difference with initial expectations.

Interacting Social robots have also been commercially used in industrial setups and other environment such as restaurants and shops. In such study Kanda et. al [21] used Robovie in a shopping mall. The Robovie had a set of behaviours which is particularly used in a shopping mall environment. Robovie was able to identify users with RFID tags, was able to carry on simple conversation, offering directions and also was able to some services of shops. Gockley et al. [22] developed and programmed Valerie, which is receptionist robot. They studies the engagement behavior of the robot and carried out a long term study for 9 months. They reported in their experiment that with the passage of time the people interaction and engagement with the robot decreased, they further proposed many design changes to make the robot more engaging for a longer period of time. Many other commercial robots have also been developed and programmed for different environments such as Robotdalen's RobCab [23] and the YDreams Siga robots [24]. The RoboCab has been developed for transportation purposes in a hospital environment while the siga robots are used for guiding purposes of visitors.

With the advancement of technology the vehicles now a days are becoming more autonomous. various companies such as Tesla are incorporated auto functionalities in their cars. Looking at the progress, probably at the end of the decade we will see completely autonomous cars on the road. With the advent of these autonomous cars it will become imperative to incorporate robots which also replaces the police personnel and current digital signals system at intersections. In a study by friedrich et. al [25] it is reported that autonomous vehicles that maneuver on roads without in the future will no longer require humans as supervisors. That is why it is necessary to understand the characteristics of traffic flow and the current system in which many elements are interdependent. They studied the on intersections with traffic signals and proposed that autonomous vehicles need to be able to communicate among themselves and with the infrastructure. Miculescu et al.

[26] proposed a a signal less coordination control algorithm for vehicles. They claim that their algorithm has provable guarantees of both safety and performance. In contrasts to this view, Sharon et .al [27] argues that as the transition to autonomous cars cannot happen abruptly in a single day, that is why a hybrid traffic managing approach that can manage both traditional vehicles and autonomous vehicles is necessary.

Building on top of that we propose that introducing a robot traffic controller at intersection will help to keep the traffic flow in more organized way and will help to avoid congestion. Gong et .al [28] in one such study developed a robot that have some hand gestures of police. These gestures are Stop, Pullover and Turn left signals. Their study rather than focusing on social behavior and engagement of the robot is more focused achieving real time performance of the robot. In another study zhao et .al [29] demonstrated traffic police gestures using 3d printed humanoid robotic arm. This study is focused on the design and analysis of robotic arm.

In our study we are more focused on social behavior of humanoid traffic police which uses several signals to direct the traffic on a busy intersection. Our focus was on how human perceive those traffic signals and how the appearance and design of the robot affect the people understanding of the gesture and the robot.

### III. STRUCTURE AND APPEARANCE OF TRAPROB

#### A. Design of TRAPROB

In this study we used Thiago++ robot [30] developed by Pal Robotics shown in Fig. 1. Thiago++ is a humanoid robot with two hands, head, trunk and wheels to move around. The Sec. (III-A1) explains in detail about the appearance and physique of modified Thiago++ robot i.e TRAPROB used in our simulation while Sec. (III-A2) explains the technical capabilities of the Thiago++.

*1) Appearance and physique of TRAPROB:* The TRAPROB is modified version of Thiago++ shown in Fig. 2 . In this version we changed the color of the robot body and also additionally put a police cap like structure on its head to mimic a traffic police. We colored the upper body as green to give shiny look just like the jacket wore by

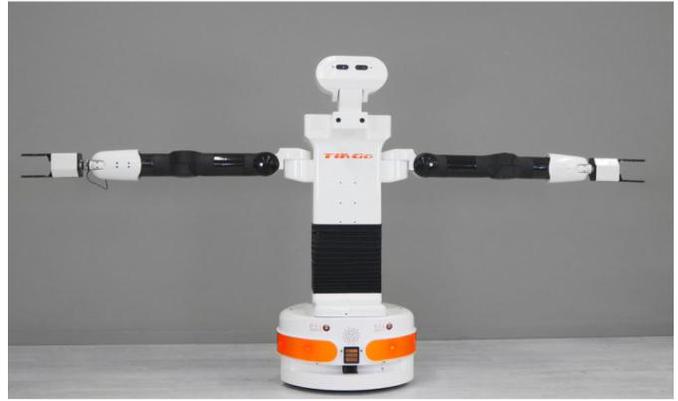

Fig. 1: Thiago++ robot

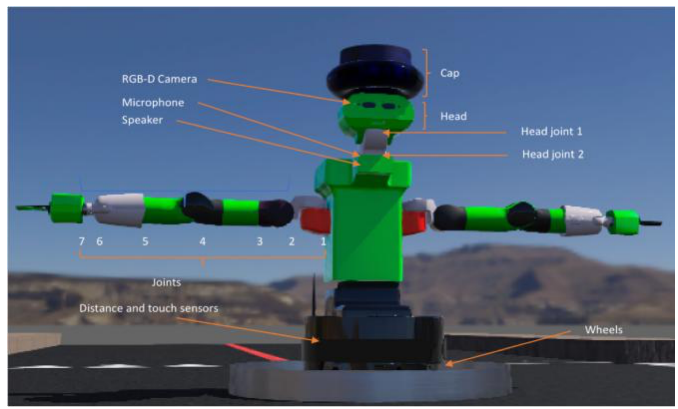

Fig. 2: TRAPBOT appearance

police personal or emergency response teams. The arms are colored with alternating green, black and white strips so that to increase the visibilty of the TRAPROB from a distance. The opposite side of shoulder is coloured red. The lower body is coloured black to imitate the pant or lower body of police personnel. The cap color is similar to that of the lower body. The

*2) Technical functionalities of the robot:* Thiago++ is a humanoid robot with height of 110cm to 145cm and has weight of 85kg. The height can be modified by stretching or curtailing the trunk which can be controlled by a motor. it can it is dual armed robot with arms having 7 degree of freedom. Each hand has seven joints and each joint has 180 degrees of freedom. The top head mounted has two joints which moves the head either in top-down fashion or left-right fashion. On the head two cameras are installed which works as an eyes of the robot. The robot has two differential wheels which which it can move around the space. The lower base also contain one touch sensor and three distance sensor which can be used to sense the surrounding place. Fig. 2 shows the detailed of all the body parts.

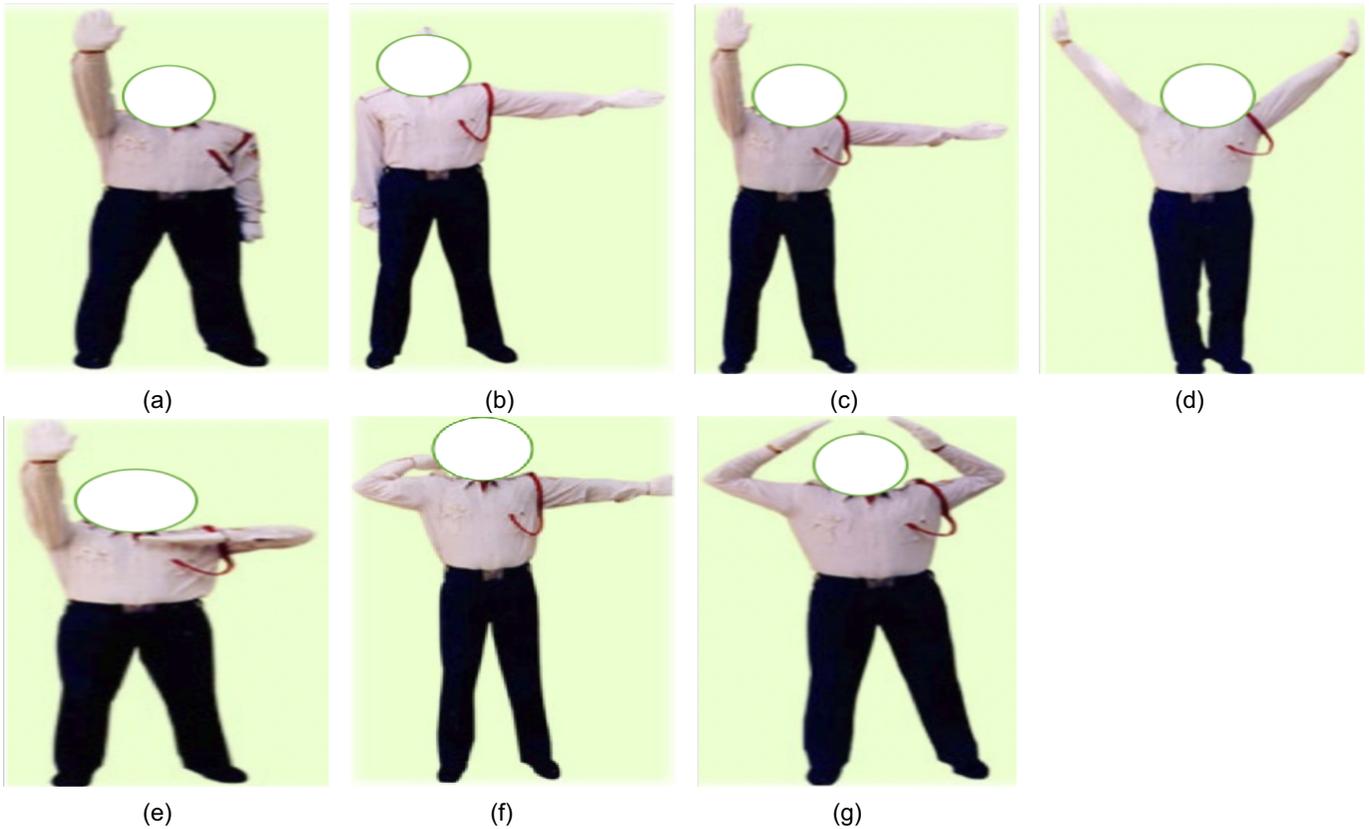

Fig. 3: Traffic police signals at intersection: (a) Front vehicles stop (b) Behind vehicles stop (c) Front and behind vehicles stop (d) Left and right vehicles stop (e) Left vehicles go (f) Right vehicles go (g) Signal change sign

## IV. TRAFFIC POLICE GESTURES

The traffic police makes different gestures to control the traffic coming from multiple ways in an intersection. These gestures are mostly understand globally but certain gestures can also be locally adopted in a country and can be varied. We will deal with the gestures described as globally accepted gestures to limit our discussion only to the social behavior of the robot. Traffic police officer position their hands to convey the signals to the drivers, driving the vehicles. Along with hand signals they also move their head and gaze to get attention of the drivers. To get a clear understanding of those signals the drivers along with their driving training also undergo through lessons in which the meaning of these signals are taught to them. We will mainly discuss 7 types of hands signals [31].

1) Front stop: To stop the approaching vehicles from the front, police personnel signal with both arm in such a way that the right arm of the police go up straight and the left hand go downward straight. The gesture is shown in Fig. Fig. 3(a)

2) Behind stop: To stop the approaching vehicles from behind, the right arm of the police personnel go down straight and the left arm stretches the left direction. The gesture is shown in Fig. 3(b)

3) Front and Behind stop: To stop the approaching vehicles from the front as well as behind, the right arm of the police go up straight and the left arm stretches the left direction. The gesture is shown in Fig. 3(c)

4) Left and right stop: To stop the approaching vehicles from the left as well as right direction, both the left and right arm of the police go up making an inclined plane. while the palms faces towards left and right respectively. The gesture is shown in Fig. 3(d)

5) Left Go: To start the vehicles from the left, police personnel signal with both hand in such a way that the right arm go up straight with the palm facing in forward direction and left arm of the police bends in the way that elbow faces leftward and fingers faces rightward. In this signal the head and also turns towards left direction. The gesture is shown in Fig. 3(e)

6) Right Go: To start the vehicles from the left, the right arm of the police bends in the way that elbow faces right direction and fingers faces left direction while the left hand is stretched in the left direction. In this signal the head and also turns towards right direction.The gesture is shown in Fig. 3(f)

7) Change sign: To signal the drivers that the sign is going to change the police officer make triangle like structure in which the left elbow points left direction, right elbow points right direction and fingers of both hands come near to each other above the head. The gesture is shown in Fig. 3(g)

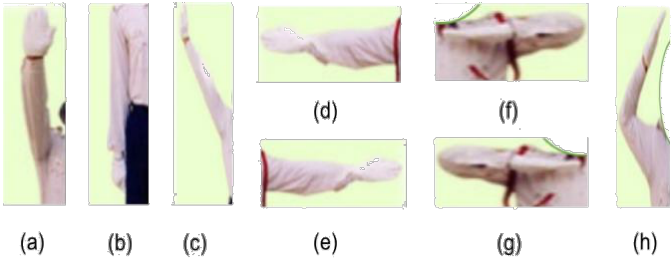

Fig. 4: Arm movement

## V. SIMULATION ENVIRONMENT AND TRAPROB GESTURES

### A. Simulation Environment

We selected Webot simulator [32] for TRAPSBOT simu-lation. The Webot package include different types of builtin robots. The thiago robot was selected for the TRAPSBOT because of its height and the length of its arms and 7 DOF movement of arms. Out of many robots developed by PAl [33], the thiago++ robots suits well with our needs as it has two robotic arms.

### B. TRAPROB Gestures creation

To embed gestures into the robot we first looked into details of the arms movements of traffic police officer. We found that arms has some basic movements out of which the signals for the traffic is created. The arm movements are shown in Fig. 4 and are classified as (a) straight up , (b) straight down, (c) inclined up, (d) leftward stretch, (e) rightward stretch, (f) left plan bent, (g) right plan bent (h) half up. These movements can be done by either hand. When movement of both these hands combines they constitute a traffic signal. In the simulator we first created these hands movements independently of each other and then combined various movements of both hand to represent a traffic signal. Our simulation currently follows wizard of oz method, and a remote human controller needs to control the robot. For a certain signal the human controller needs to press a button for signalling the traffic. The Fig . 6 shows the overall flowchart of the signals creation. The red color of the box represent stopping gestures, the red signalize go, the yellow is the interim signal between different signal, the left blue boxes are the left arm movement, the right blue boxes are the right arm movements, the left black box are the left arm joints affected by the movement and the right black boxes are the right arm joints affected by the movement. Final gestures by the TRAPSBOT based on the flow chart is shown is shown in Fig. 7

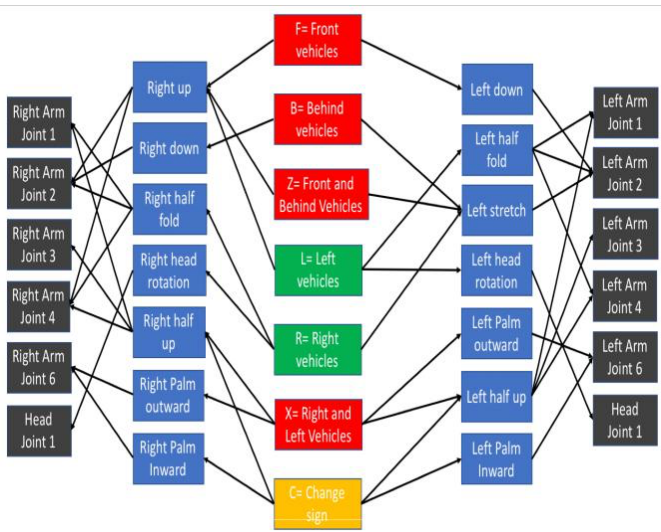

Fig. 5: The gesture making flow chart

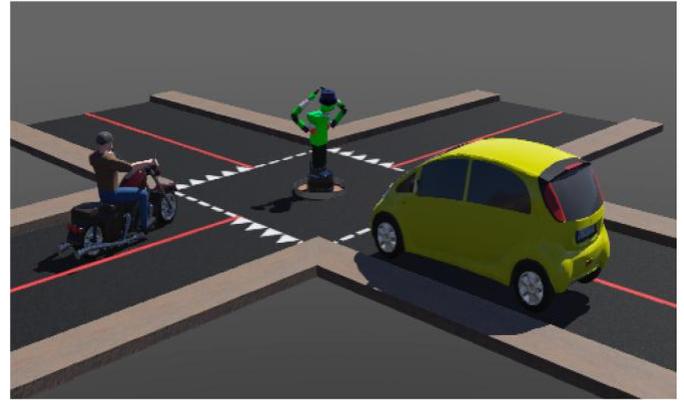

Fig. 6: The Webot traffic controlling environment

### C. TRAPROB road traffic environment

To give a real look road traffic environment look we then added a four way intersection and added vehicles to the environment. The road environment is necessary as we want to know about how the user perceive the environment and whether they can identify the robot from its appearance as a traffic robot or not. The whole environment is shown in Fig.

## VI. EXPERIMENTAL SETUP

Through this experiment we will evaluate the social behavior of robot and how people perceive the gesture of the robots. Physical evaluation of the robot by installing it at intersection for time span is beyond the scope of this study, that's why we choose to evaluate the robot behavior solely based on simulation. In a normal traffic environment at intersection three group of people are involved that can give their opinion and evaluate the scenarios. These group of people are:

1) The drivers that drive through the intersection.
2) The pedestrians that walk through the intersection.
3) The traffic police officer controlling the traffic.

We only evaluated the simulation from drivers perspectives. We didn't evaluated it based on pedestrian perspective of the robot as our gestures only included traffic signals that direct the vehicles. As access to police to police personal were difficult in this time, we also skipped the evaluation from traffic police perspective.

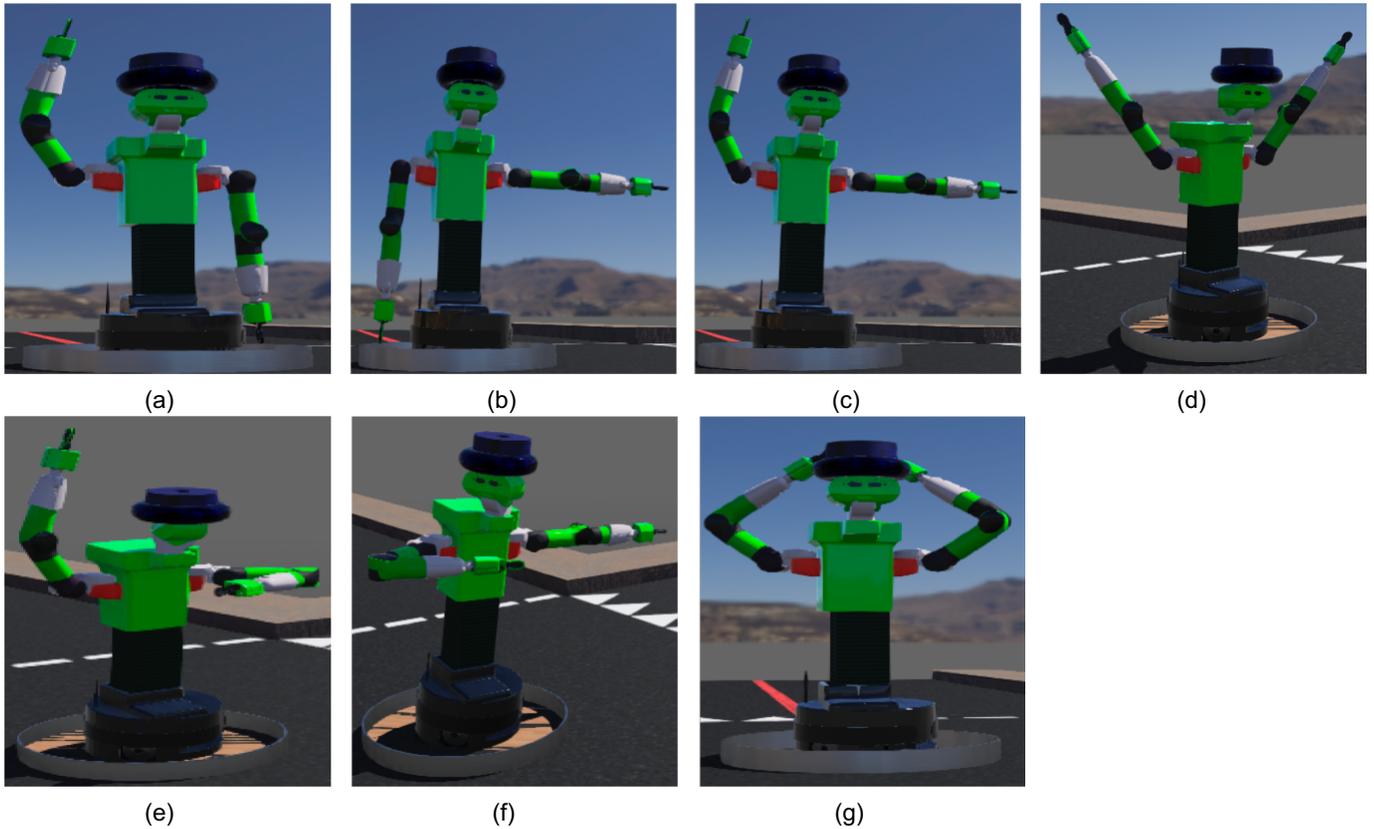

Fig. 7: Traffic robot police signals at intersection: (a) Front vehicles stop (b) Behind vehicles stop (c) Front and behind vehicles stop (d) Left and right vehicles stop (e) Left vehicles go (f) Right vehicles go (g) Signal change sign

To evaluate the social behavior of the robot a huge pop-ulation is needed but due to limitations of the study, time constraint the simulation was evaluated by two colleagues. The participants were briefed before the study and a written consent was acquired. We classified the TRAPBOT evaluation into three major parts.

### A. The TRAPBOT simulation environment

This experiment was designed to evaluate whether people can identify whether it is traffic robot or not. The partici-pant were provided three versions of the simulation. In the first version a simple thiago++ robot was shown without any modified appearance and any surrounding environment. In the second version we show then the TRAPBOT with modified appearance. In the third version, the participants were introduced to the TRAPBOT with the the traffic intersection environment. In all three versions the participants were shown random gestures. In all three version same options were given to the participant to identify the robot. The options given were waving robot, dancing robot, traffic controlling robot, general robot, balloon mascot and none of them.

### B. The appearance of the TRAPBOT.

This experiment was designed to know that whether the appearance which we have given to the TRAPBOT is similar to the traffic police personnel or not. The participants were asked about the color combination, the height, arms length and the police hat. A Likert scale questionnaire was prepared for this purpose and its shown in table I.

### C. The gestures of the TRAPBOT.

To understand whether driver will be able to decode the gestures of the robot as they decode the gestures of police officers we asked participants about various random arm ges-tures and their understanding of the gestures. The participant knowledge of the traffic sign were first evaluated by showing them pictures of police officer making hand gestures similar to those shown in figure 7. After recording the knowledge level the participants about the traffic police gestures we then show them the gestures made by the TRAPSBOT and asked them to identify the gesture. The gestures were shown in random way and some of the gestures were even repeated. The participants were then given another Likert scale questionnaire to find the whether the arms motion, and the movement resembles the a traffic police officer or in general a human movement or not. The questionnaire is shown in table II

## VII. RESULTS AND DISCUSSION

We carried out our experiments with small size and very controlled environment and the results might not represents the exact social social behavior of the TRAPBOT. Still we are positive that it gives us some in depth insights about our simulation.

Evaluate the appearance of the robot in light of the following the statements.

| | Strongly Disagree | Disagree | Neutral | Agree | Strongly Agree |
|---|---|---|---|---|---|
| The height of the robot well suits the environment | | | | | |
| The arms are clearly visible in the simulation setup. | # | # | # | # | # |
| The arm's length is perfect for traffic signals at the intersection. | # | # | # | # | # |
| The robot appearance looks similar to the police officer. | # | # | # | # | # |
| The green color is catchy and can be noticed at intersection. | # | # | # | # | # |
| The hat wore by the robot gives it a police officer look. | # | # | # | # | # |
| The green, black and white strips on robots make it more visible. | # | # | # | # | # |
| The robot overall looks like a police officer. | # | # | # | # | # |
| | # | # | # | # | # |

TABLE I: TRAPBOT Appearance Evaluation Questionnaire

Evaluate the appearance of the robot in light of the following the statements.

| | Strongly Disagree | Disagree | Neutral | Agree | Strongly Agree |
|---|---|---|---|---|---|
| The arms movement of the robot resemble a human being. | | | | | |
| The speed of arms movement is fine for traffic control. | # | # | # | # | # |
| The gestures made by the robot is readable by the drivers when seen. | # | # | # | # | # |
| The robot do not have any risk of falling down while moving arms. | # | # | # | # | # |
| The robot gestures are scary an shouldn't be on the intersection. | # | # | # | # | # |
| The robot movements resemble the traffic police officer. | # | # | # | # | # |
| The gestures made for different signals vary from each other. | # | # | # | # | # |
| The head movement is synchronized with arms movement. | # | # | # | # | # |
| The head movement is similar to that of police officer. | # | # | # | # | # |
| | # | # | # | # | # |

TABLE II: TRAPBOT Movement Evaluation Questionnaire

| | Thaigo++ | TRAPSBOT | TRAPSBOT at intersection |
|---|---|---|---|
| P1 | DR | TCR | TCR |
| P2 | DR | DR | TCR |
| ID ratio | 0% | 50% | 100% |

P=participant, DR=Dancing Robot, TCR=Traffic controlling robot ID ratio= percentage of person who identified TRAPBOT

TABLE III: TRAPBOT Identification Evaluation

In the first experiment of simulation environment the questions we were looking for were how the people see the robot and whether they can identify that robot was designed to control the traffic or not. Our hypothesis was that robot when installed at traffic intersection will be identified as traffic robot. The results are summarized in table III. Both the participants were able to recognize the TRAPBOT as traffic controlling robot, when installed at traffic intersection. Only half of the participant identified it as traffic robot when standalone TRAPBOT was shown. None of participants identified thiago++ robot as traffic controlling robot even it had traffic gestures.

The appearance of the robot was analyzed through by asking participants to answer a Likert scale questionnaire. The responses are shown in Fig. 8. The average response of the user mostly lie in the upper scale, i.e they strongly agree that robot appearance in all aspects resembles the the police officer. The participants on average both disagree that the hat which the TAPSBOT wear does not resembles with traffic police hat and should be modified. More the standard deviation between the responses of both the participants is below than 1 and their answers lie quite close to each other, except for the arm length where the standard deviation is higher than other questions.

The robot gestures and social behavior were evaluated by the participants by looking at various gestures. The responses are shown in Fig 9. The participants rated the TRAPBOT as having clear and readable gestures gestures. Their responses shows that robot has high socially interactive and functionality of the robot is good enough to be installed at traffic intersection. The standard deviation between the responses of the participant is very low and they have given the same response to most of the questions. Both the participants agree that gestures are clear, readable, and different from each other, its similar to the gestures made by the police officer and the robot movement is more human like. The only thing which they don't like is that the speed of movement of arms. Upon further investigation from participants in discussion, they revealed that the speed of the arms looks more dramatic and should be increased to save time and look more real.

Looking at above results, although both the participants agree on most of the questions we cannot take a firm stand to whether the TRAPBOT is feasible to be installed at traffic intersection, but we will definitely recommend to install it for further evaluation. More over the appearance of robots needs more modifications and designers can be asked to modify the look and appearance of TRAPBOT so that it look more like a traffic police. Also, In our study we only asked a generalized questions about the gestures but we haven't evaluated each gesture alone to find out whether participants can recognize the signal are not. This experiment will reveal more in depth behavior of the robot and each gesture will be modified to resemble the gestures of police officer.

## VIII. CONCLUSION

The police officer experiences high stress because of working for high number of hours. To replace the human we need to design a robot which is socially acceptable, which can have interactive social behavior, whose appearance look like a police officer and which can offer the same gestures functionality as a police officer have. In this we study we designed one such traffic police robot (TRAPBOT) using the Webot environment and then evaluated its social behavior.

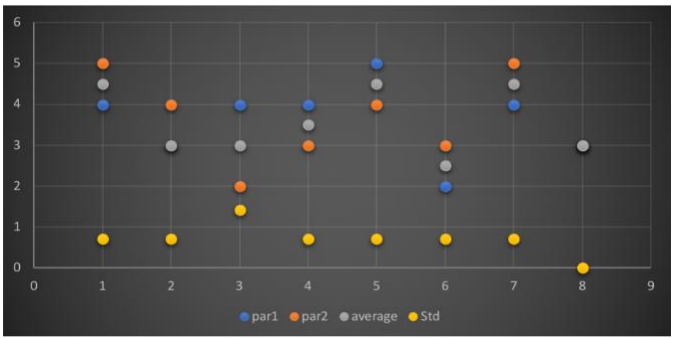

Fig. 8: TRAPBOT appearance analysis

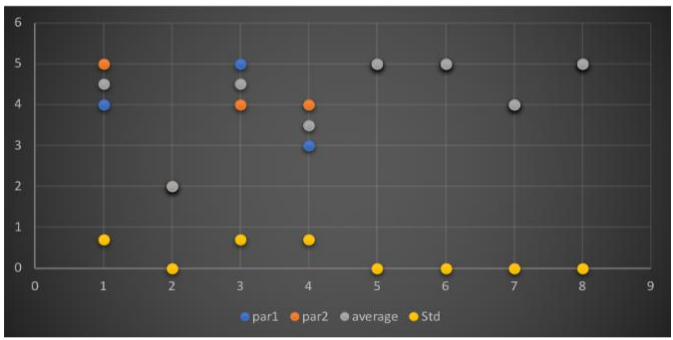

Fig. 9: TRAPBOT social behavior analysis

Two participants were recruited to evaluate the TRAPBOT. We concluded in our study that though the participants have rated the robot as social and interactive, we still need higher number of population to evaluate the performance of a social behavior of TRAPBOT. More ever in the future work we need to simulate an environment in which the vehicles moves in response to the robot signals. To deploy the TRAPBOT a long time span study of physical robot at an intersection will also be carried out to validate the the outcomes of the simulation study.

## APPENDIX A
### TRAFFIC ROBOT CONTROLLER

```python
from controller import Robot,Keyboard ,Motor, DistanceSensor
import math
class thiago (Robot):
    #create basic hand signals
    def right_up(self):
        self.right_half_arm.setPosition(-1.11)
        self.right_arm.setPosition(0.7)
        self.right_shoulder.setPosition(0)
        self.right_half_hand.setPosition(0)
        self.right_hand.setPosition(0)
        self.right_mid_arm.setPosition(0)
        self.head.setPosition(0)

    def left_up(self):
        self.left_half_arm.setPosition(-1.11)
        self.left_arm.setPosition(0.7)
        self.left_shoulder.setPosition(0)
        self.left_half_hand.setPosition(0)
        self.left_hand.setPosition(0)
        self.left_mid_arm.setPosition(0)
        self.head.setPosition(0)

    def right_down(self):
        self.right_half_arm.setPosition(1.5)
        self.right_arm.setPosition(0)
        self.right_shoulder.setPosition(0)
        self.right_half_hand.setPosition(0)
        self.right_hand.setPosition(0)
        self.right_mid_arm.setPosition(0)
        self.head.setPosition(0)

    def left_down(self):
        self.left_half_arm.setPosition(1.5)
        self.left_arm.setPosition(0)
        self.left_shoulder.setPosition(0)
        self.left_half_hand.setPosition(0)
        self.left_hand.setPosition(0)
        self.left_mid_arm.setPosition(0)
        self.head.setPosition(0)

    def right_half(self):
        self.right_shoulder.setPosition(0.5)
        self.right_arm.setPosition(2.29)
        self.right_mid_arm.setPosition(1.5)
        self.right_half_arm.setPosition(0)
        self.right_half_hand.setPosition(0)
        self.right_hand.setPosition(0)
        self.head.setPosition(0)

    def left_half(self):
        self.left_shoulder.setPosition(0.5)
        self.left_arm.setPosition(2.29)
        self.left_mid_arm.setPosition(1.5)
        self.left_half_arm.setPosition(0)
        self.left_half_hand.setPosition(0)
        self.left_hand.setPosition(0)
        self.head.setPosition(0)

    def left_straight(self):
        self.left_shoulder.setPosition(-0.5)
        self.left_arm.setPosition(0)
        self.left_mid_arm.setPosition(0)
        self.left_half_arm.setPosition(0)
        self.left_half_hand.setPosition(0)
        self.left_hand.setPosition(0)
        self.head.setPosition(0)

    def right_straight(self):
        self.right_shoulder.setPosition(0.5)
        self.right_arm.setPosition(0)
        self.right_mid_arm.setPosition(0)
        self.right_half_arm.setPosition(0)
        self.right_half_hand.setPosition(0)
        self.right_hand.setPosition(0)
        self.head.setPosition(0)

    def left_half_up(self):
        self.left_shoulder.setPosition(0.5)
        self.left_half_arm.setPosition(-1)
        self.left_arm.setPosition(0)
        self.left_half_hand.setPosition(-1.5)
        self.left_mid_arm.setPosition(0)
        self.left_hand.setPosition(0)
        self.head.setPosition(0)

    def right_half_up(self):
        self.right_shoulder.setPosition(-0.5)
        self.right_half_arm.setPosition(-1)
        self.right_arm.setPosition(0)
        self.right_half_hand.setPosition(-1.5)
        self.right_mid_arm.setPosition(0)
        self.right_hand.setPosition(0)
        self.head.setPosition(0)

    def left_half_fold(self):
        self.left_shoulder.setPosition(0.5)
        self.left_half_arm.setPosition(-0.7)
        self.left_arm.setPosition(1.8)
        self.left_half_hand.setPosition(-1.5)
        self.left_mid_arm.setPosition(0)
        self.left_hand.setPosition(0)
```

```python
    def right_half_fold(self):
        self.right_shoulder.setPosition(-0.5)
        self.right_half_arm.setPosition(-0.7)
        self.right_arm.setPosition(1.8)
        self.right_half_hand.setPosition(-1.5)
        self.right_mid_arm.setPosition(0)
        self.right_hand.setPosition(0)

    def look_left(self):
        self.head.setPosition(1.1)

    def look_right(self):
        self.head.setPosition(-1.1)

    #create traffic signals by combining hand signals
    def front_stop(self):
        self.right_up()
        self.left_down()

    def behind_stop(self):
        self.right_down()
        self.left_straight()

    def front_behind_stop(self):
        self.right_up()
        self.left_straight()

    def left_right_stop(self):
        self.left_half_up()
        self.right_half_up()

    def all_stop(self):
        self.right_up()
        self.left_up()

    def start_left(self):
        self.right_up()
        self.left_half()
        self.look_left()

    def start_right(self):
        self.right_half()
        self.left_straight()
        self.look_right()

    def change_sign(self):
        self.right_half_fold()
        self.left_half_fold()

    def activate(self):
        self.timestep = int(self.getBasicTimeStep())
        self.left_shoulder = self.getDevice('arm left 1 joint')
        self.left_half_arm = self.getDevice('arm left 2 joint')
        self.left_mid_arm = self.getDevice('arm left 3 joint')
        self.left_arm = self.getDevice('arm left 4 joint')
        self.left_half_hand=self.getDevice('arm left 5 joint')
        self.left_hand=self.getDevice('arm left 6 joint')
        self.right_shoulder = self.getDevice('arm_right 1 joint')
        self.right_half_arm = self.getDevice('arm right 2 joint')
        self.right_mid_arm = self.getDevice('arm right 3 joint')
        self.right_arm = self.getDevice('arm right 4 joint')
        self.right_half_hand=self.getDevice('arm right 5 joint')
        self.right_hand=self.getDevice('arm right 6 joint')
        self.tall=self.getDevice('torso lift joint')
        self.tall.setPosition(0.35)
        self.head=self.getDevice('head 1 joint')
        # self.keyboard = self.getKeyboard()
        # self.keyboard.enable(10 * self.timestep)

    def __init__(self):
        Robot.__init__(self)
        self.activate()

    def run(self):
      while robot.step(self.timestep) != -1:
            # Enter here functions to send actuator com-mands, like:
            key = self.keyboard.getKey()
            # if key == Keyboard.UP:
                    # self.right_straight_up()
            # if key== Keyboard.DOWN:
                    # self.right_straight_down()
            # if key== Keyboard.RIGHT:
                    # self.left_straight_down()
            # if key== Keyboard.LEFT:
                    # self.left_straight_up()
            if key== ord('F'):
                    self.front_stop()
            if key== ord('B'):
                    self.behind_stop()
            if key== ord('Z'):
                    self.front_behind_stop()
            if key== ord('X'):
                    self.left_right_stop()
            if key== ord('L'):
                    self.start_left()
            if key== ord('R'):
                    self.start_right()
                    self.look_right()
            if key==ord('C'):
                    self.change_sign()
            pass
robot = thiago()
robot.run()
```